\documentclass[sigconf]{acmart}

\renewcommand\footnotetextcopyrightpermission[1]{} % removes footnote with conference information in first column
\usepackage{hyperref}

\def\BibTeX{{\rm B\kern-.05em{\sc i\kern-.025em b}\kern-.08emT\kern-.1667em\lower.7ex\hbox{E}\kern-.125emX}}
    
\copyrightyear{2018}
\acmYear{2018}
\setcopyright{acmlicensed}
\acmConference[KDD '19]{xxx}{June 03--05, 2019}{Anchorage, AK}
\acmBooktitle{KDD '19: ACM Symposium on Neural Gaze Detection, June 03--05, 2018, Woodstock, NY}
\acmPrice{15.00}
\acmDOI{10.1145/1122445.1122456}
\acmISBN{978-1-4503-9999-9/18/06}

\acmSubmissionID{123-A56-BU3}

\begin{document}

%
% The "title" command has an optional parameter, allowing the author to define a "short title" to be used in page headers.
\title{Efficient Pipeline for Camera Trap Image Review}

%
% The "author" command and its associated commands are used to define the authors and their affiliations.
% Of note is the shared affiliation of the first two authors, and the "authornote" and "authornotemark" commands
% used to denote shared contribution to the research.
\author{Sara Beery}
\email{sbeery@caltech.edu}
\affiliation{%
  \institution{California Institute of Technology}
  \streetaddress{1200 E California Blvd.}
  \city{Pasadena}
  \state{California}
  \postcode{91125}
}

\author{Dan Morris}
\email{dan@microsoft.com}
\affiliation{%
  \institution{Microsoft AI for Earth}
  \streetaddress{14820 NE 36th Street}
  \city{Redmond}
  \state{Washington}
  \postcode{98052}
}

\author{Siyu Yang}
\email{yasiyu@microsoft.com}
\affiliation{%
  \institution{Microsoft AI for Earth}
  \streetaddress{14820 NE 36th Street}
  \city{Redmond}
  \state{Washington}
  \postcode{98052}
}

%
% By default, the full list of authors will be used in the page headers. Often, this list is too long, and will overlap
% other information printed in the page headers. This command allows the author to define a more concise list
% of authors' names for this purpose.

%
% The abstract is a short summary of the work to be presented in the article.
\begin{abstract}
Biologists all over the world use camera traps to monitor biodiversity and wildlife population density. The computer vision community has been making strides towards automating the species classification challenge in camera traps~\cite{ren2013ensemble,yu2013automated,wilber2013animal,chen2014deep,lin2014foreground,swanson2015snapshot,zhang2015coupled,zhang2016animal,miguel2016finding,giraldo2017camera,yousif2017fast,villa2017towards,norouzzadeh2018automatically, beery2018recognition, beery2019synthetic}, but it has proven difficult to to apply models trained in one region to images collected in different geographic areas. In some cases, accuracy falls off catastrophically in new region,  due to both changes in background and the presence of previously-unseen species.  We propose a pipeline that takes advantage of a pre-trained general animal detector and a smaller set of labeled images to train a classification model that can efficiently achieve accurate results in a new region.
\end{abstract}


\ccsdesc{Computing methodologies~Machine learning}

%
% Keywords. The author(s) should pick words that accurately describe the work being
% presented. Separate the keywords with commas.
\keywords{detection, classification, camera traps, biodiversity monitoring}

%
% A "teaser" image appears between the author and affiliation information and the body 
% of the document, and typically spans the page. 
\begin{teaserfigure}
\centering
  \includegraphics[width=\textwidth]{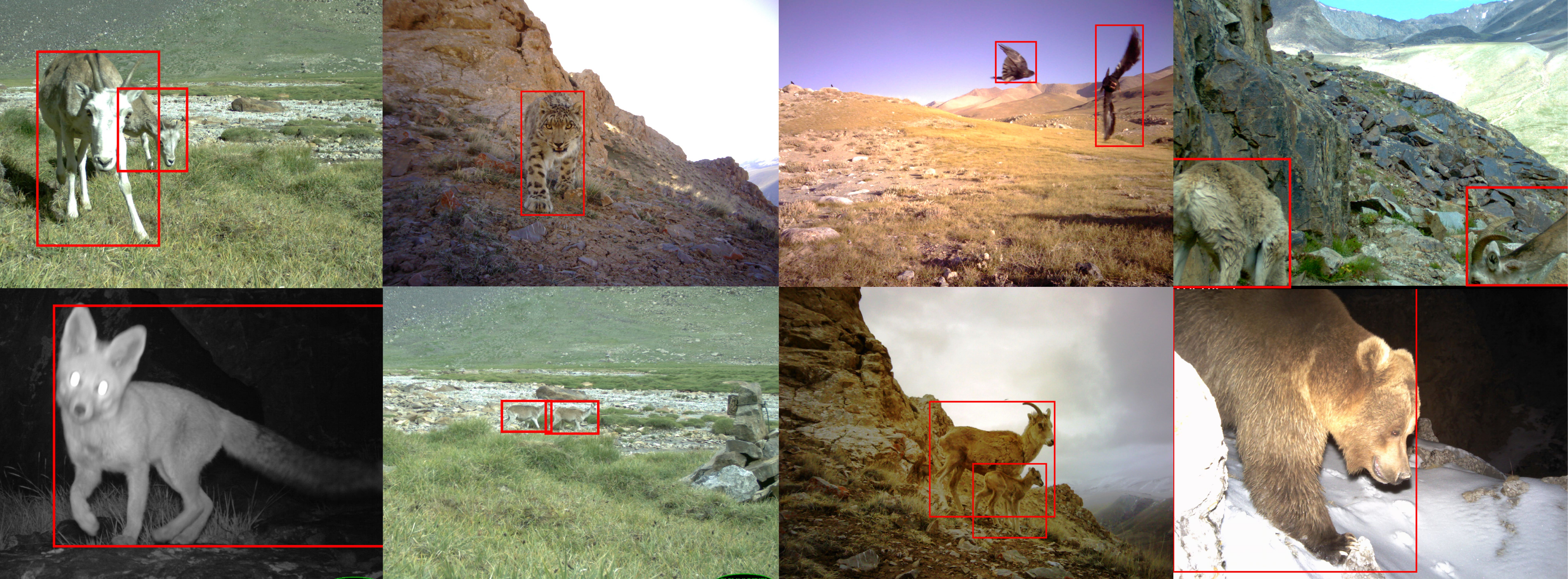}
  \caption{Example results from our generic detector, on images from regions and/or species not seen during training.}
  \Description{Results from our generic detector, on images from regions and/or species not seen during training.}
  \label{fig:teaser}
\end{teaserfigure}

%
% This command processes the author and affiliation and title information and builds
% the first part of the formatted document.
\maketitle

\section{Introduction}
Camera traps are heat- or motion-activated cameras placed in the wild to monitor and investigate animal populations and behavior. They are used to locate threatened species, identify important habitats, monitor sites of interest, and analyze wildlife activity patterns. At present, the time required to manually review images severely limits productivity. Additionally, \textasciitilde70\% of camera trap images are empty, due to a high rate of false triggers.

Previous work has shown good results on automated species classification in camera trap data \cite{norouzzadeh2018automatically}, but further analysis has shown that these results do not generalize to new cameras or new geographical regions~\cite{Beery_2018_ECCV}. Additionally, these models will fail to recognize any species they were not trained on. In theory, it is possible to re-train an existing model in order to add missing species, but in practice, this is quite difficult and requires just as much machine learning expertise as training models from scratch. Consequently, very few organizations have successfully deployed machine learning tools for accelerating camera trap image annotation.

We propose a different approach to applying machine learning to camera trap projects, combining a \textit{generalizable detector} with \textit{project-specific classifiers}.

We have trained an animal detector that is able to find and localize (but not identify) animals, even species not seen during training, in diverse ecosystems worldwide.  See Fig.~\ref{fig:teaser} for examples of the detector on camera trap images from regions and/or species not seen during training. By first finding and localizing animals, we are able to:
\begin{enumerate}
    \item{drastically reduce the time spent filtering empty images, and}
    \item{dramatically simplify the process of training species classifiers, because we can crop images to individual animals (and thus classifiers need only worry about animal pixels, not background pixels).}
\end{enumerate}
With this detector model as a powerful new tool, we have established a modular pipeline for on-boarding new organizations and building project-specific image processing systems. 
\section{Pipeline}
We break our pipeline into four stages: data ingestion, animal detection, classifier training, and application to new data.

\subsection{Data ingestion}
 
First we transfer images to the cloud, either by uploading to a drop point or by mailing an external hard drive.  Data comes in a variety of formats; we convert each dataset to the COCO-Camera Traps format, i.e., we create a JSON file that encodes the annotations and the image locations within the organization's file structure.
 
\subsection{Animal detection}
 
We next run our (generic) animal detector on all the images to locate animals. We have developed an infrastructure for efficiently running this detector on millions of images, dividing the load over multiple nodes.
 
We find that a single detector works for a broad range of regions and species. If the detection results (as validated by the organization) are not sufficiently accurate, it is possible to collect annotations for a small set of their images and fine-tune the detector. Typically these annotations would be fed back into a new version of the general detector, improving results for subsequent projects.
 
\subsection{Classifier training}
 
Using species labels provided by the organization, we train a (project-specific) classifier on the cropped-out animals.
 
\subsection{Application to new data}
 
We use the general detector and the project-specific classifier to power tools facilitating accelerated verification and image review, e.g., visualizing the detections, selecting images for review based on model confidence, etc.
\section{Case Study: The Idaho Department of Fish and Game}
We applied our pipeline to 4.8 million images collected by the Idaho Department of Fish and Game (IDFG) from six regions in the state, of which 0.76 million have image-level species labels. Spreading the load over 16 nodes, each with one GPU, it took under three days to perform detection on this batch of images. By filtering out images without confident detections, we have eliminated some 80\% of images (estimated by the project owner at IDFG) from manual review as this study contained a large percentage of empty frames. The average precision for animal detections ranges from 0.885 to 0.988 for different regions, evaluated against species labels as an indication of animal presence.

Notably, the version of the detector used was not trained with any camera trap images with snow but performed very well on such images in the IDFG data. The detector was also able to find animals in night images that reviewers would have missed without adjusting the exposure of the image. False positives were a problem where branches and rocks were misidentified as animals. A post-processing step to remove detections that appear in the same position for many frames in a row alleviated this issue. We are in the process of training the project-specific classifier for IDFG, and preliminary results for species classification are promising.
\section{Conclusions}
We propose a pipeline that allows us to train classifiers for new camera trap projects in an efficient way, first leveraging a generic animal detection model to localize animals and remove empties, then training a project-specific classifier using the localized images of animals and their image-level labels. We present this as a new approach to structuring camera trap projects, and aim to formalize discussion around the steps that are required to successfully apply machine learning to camera trap images. 

Our code and models are available at \href{http://github.com/microsoft/cameratraps}{github.com/microsoft/cameratraps}, and public datasets used for training are available at \href{http://lila.science}{lila.science}.

\bibliographystyle{ACM-Reference-Format}
\bibliography{main}
\end{document}